\documentclass{article}
\usepackage{spconf,amsmath}
\usepackage{multirow}
\usepackage{graphicx}
\usepackage{booktabs}
\usepackage{tabularx}
\renewcommand{\paragraph}[1]{\vspace{1em}\noindent\textbf{#1}~~}
\usepackage{amsmath,amssymb}
\usepackage{listings}
\usepackage{wrapfig}
\usepackage{url}
\usepackage{caption}
\DeclareCaptionFont{ninehalf}{\fontsize{9.5}{11.4}\selectfont}
\title{E{\small asy}C{\small ontrol}E{\small dge}: A Foundation-Model Fine-Tuning for \\ Edge Detection}
\name{Hiroki Nakamura\textsuperscript{1}, Hiroto Iino\textsuperscript{1}, Masashi Okada\textsuperscript{1}, Tadahiro Taniguchi\textsuperscript{2}}
\address{\textsuperscript{1}Panasonic Holdings Corporation, \textsuperscript{2}Kyoto University}
\begin{document}
%
\maketitle
\begin{abstract}
  We propose \textbf{EasyControlEdge}, adapting an image-generation foundation model to edge detection.
  In real-world edge detection (e.g., floor-plan walls, satellite roads/buildings, and medical organ boundaries), \textbf{crispness} and \textbf{data efficiency} are crucial, yet producing crisp raw edge maps with limited training samples remains challenging.
  Although image-generation foundation models perform well on many downstream tasks, their pretrained priors for data-efficient transfer and iterative refinement for high-frequency detail preservation remain underexploited for edge detection.
  To enable crisp and data-efficient edge detection using these capabilities, we introduce an edge-specialized adaptation of image-generation foundation models.
  To better specialize the foundation model for edge detection, we incorporate an edge-oriented objective with an efficient pixel-space loss.
  At inference, we introduce guidance based on unconditional dynamics, enabling a single model to control the edge density through a guidance scale.
  Experiments on BSDS500, NYUDv2, BIPED, and CubiCasa compare against state-of-the-art methods and show consistent gains, particularly under no-post-processing crispness evaluation and with limited training data.
\end{abstract}
\section{Introduction}
\label{sec:intro}
\begin{figure*}[t]
  \centering
  \includegraphics[width=0.92\textwidth]{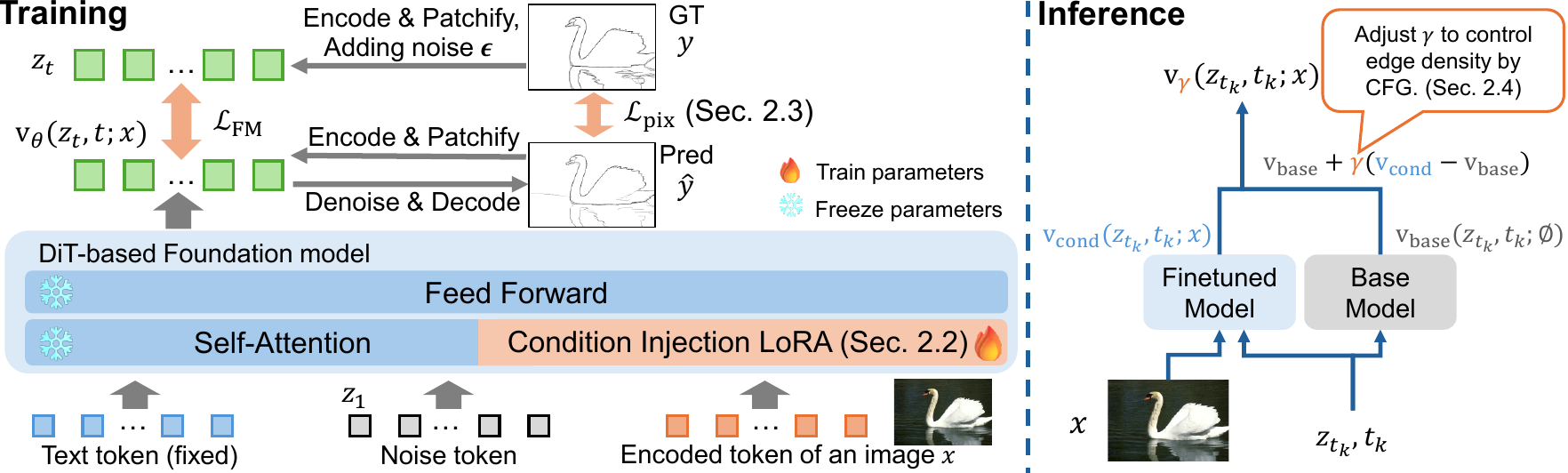}
  \caption{Overview of EasyControlEdge.
    The left side shows the training flow and the right side shows the inference flow.
    We train only a condition-injection LoRA on a frozen DiT-based foundation model (Sec.~\ref{sec:cond_inject}) with edge-oriented objectives (Sec.~\ref{sec:pix_loss}), and control edge density at inference via classifier-free guidance with scale $\gamma$ (Sec.~\ref{sec:fm_guidance}).}
  \label{fig:abstract}
\end{figure*}
Edge detection provides an informative representation of image structure by extracting boundaries and contours~\cite{canny1986}, making it a key component in a wide range of pipelines, from recognition and detection to segmentation~\cite{cheng2020boundary, huang2025comprehensive, jing2022recent}.
Edge detection is widely deployed across diverse domains, such as medical imaging, road detection in remote sensing, and wall boundary detection in floor plans~\cite{huang2025comprehensive, jing2022recent, zeng2019deep}.
In addition, edge maps often provide crucial information for downstream tasks including boundary vectorization and floor-plan reconstruction~\cite{zeng2019deep, liu2017raster}.

For practical edge detection, \textbf{crispness} and \textbf{data efficiency} are important.
\textbf{Crispness} refers to producing thin, well-localized edges without excessive reliance on post-processing~\cite{wang2017deepcrisp, huan2021unmixing, ye2023delving, ye2024diffusionedge, zhou2024notricks}.
However, much of the prior work emphasizes correctness, i.e., classifying pixels as edge/non-edge~\cite{wang2017deepcrisp},
and performance comparisons are typically conducted on benchmark datasets with hundreds or more training images~\cite{MartinFTM01,Silberman:ECCV12,soria2020dexined}.
Moreover, many benchmarks compute metrics after non-maximum suppression (NMS) and thinning, so improvements may reflect the overall pipeline rather than raw predictions~\cite{zhou2024notricks, ye2024diffusionedge}.
Modern architectures involving downsampling/upsampling often reduce spatial fidelity, yielding thicker responses and increasing reliance on post-processing, which can be fragile for closely spaced boundaries and subtle structures~\cite{wang2017deepcrisp, ye2023delving, zhou2024notricks}.
\textbf{Data efficiency} refers to achieving strong performance with limited training samples.
This is important because large-scale data collection and annotation are often costly and operationally challenging in practice.

To address crispness and data efficiency, modern image-generation foundation models~\cite{rombach2022ldm,esser2024scaling,batifol2025flux} present a promising direction: they encode rich priors from large-scale pretraining and can progressively refine outputs through iterative inference while preserving high-frequency details.
In fact, adapting such foundation models to downstream tasks via lightweight fine-tuning~\cite{hu2022lora, zhang2025easycontrol} has been shown to reduce blur and preserve fine high-frequency structures, often yielding crisper outputs than earlier task-specific approaches~\cite{xu2025ootdiffusion, qiu2025adaptively}.
For instance, EasyControl~\cite{zhang2025easycontrol} injects conditioning signals via minimal trainable parameters, efficiently adapting a foundation model to a specific task.

Although edge detection has advanced through new learning methods and architectures with deep learning~\cite{ye2024diffusionedge, zhou2024ged,pu2022edter,yang2024edgesam,li2025edmb}, existing approaches have not fully translated these foundation-model strengths into crisp and data-efficient edge detection.
GED~\cite{zhou2024ged} transfers Stable Diffusion~\cite{rombach2022ldm} priors but predicts edges in a single step, without leveraging iterative refinement.
DiffusionEdge~\cite{ye2024diffusionedge} employs iterative diffusion to obtain relatively crisp edges, but does not exploit foundation-model priors.
Therefore, achieving crisp edge generation with limited training data requires retaining the pretrained priors and iterative refinement that generate high-frequency details, and specializing them to edge detection.

In this work, we propose \textbf{EasyControlEdge}, a framework that specializes modern generative foundation models for edge detection to achieve \textbf{crispness} and \textbf{data efficiency}.
The overview of our method is illustrated in Fig.~\ref{fig:abstract}.
Our approach has three key ideas.
(1) We fine-tune an image-generation foundation model via lightweight adaptation and leverage its pretrained priors and iterative generation to generate crisp boundaries with limited annotations.
(2) We add an edge-specific pixel loss with efficient backpropagation to improve performance with limited extra computational cost.
(3) At inference time, we adopt a Classifier-Free Guidance (CFG)-based scheme~\cite{ho2021classifier, phunyaphibarn2025unconditional} that adjusts edge density via the guidance scale without retraining.

Experiments on standard edge detection benchmarks~\cite{MartinFTM01,Silberman:ECCV12,soria2020dexined} and wall-boundary detection on CubiCasa~\cite{kalervo2019cubicasa5k} show that our method is competitive with, or outperforms, prior works, with particularly large gains on CEval (raw outputs without post-processing), demonstrating sharp, well-localized boundaries.
We also show strong performance  with limited training data (e.g., $n<100$) and demonstrate controllability of edge density.
%
%
\section{Method: E{\footnotesize asy}C{\footnotesize ontrol}E{\footnotesize dge}}
\label{sec:method}
In this section, we present \textbf{EasyControlEdge}, a framework for specializing modern image-generation foundation models for edge detection.
Our key idea is to
(i) adapt a strong foundation model to edge detection via lightweight condition injection (Sec.~\ref{sec:cond_inject}),
(ii) introduce pixel-space supervision with efficient backpropagation to enforce pixel-accurate localization (Sec.~\ref{sec:pix_loss}), and
(iii) enable controllability through guidance at inference (Sec.~\ref{sec:fm_guidance}).

We build on FLUX~\cite{batifol2025flux, fluxdev} and EasyControl~\cite{zhang2025easycontrol} as our foundation and base the following descriptions on this implementation; however, we note that the proposed design choices are broadly applicable to other conditional generative models and fine-tuning methods.

\subsection{Preliminaries}
Given an input image $x$ and a target edge map $y$, we cast edge detection as conditional generation $p(y\mid x)$ in the autoencoder latent space, solving it via rectified flow matching~\cite{lipman2023flow}, where $E(\cdot)$ and $D(\cdot)$ denote the encoder and decoder, respectively.
We obtain the clean latent $\mathbf{z}_0=E(y)$ and decode a latent $\mathbf{z}$ back to an edge map by $D(\mathbf{z})$, and learn a conditional time-dependent vector field to generate $\hat{y}$ by integrating the induced ODE from $t=1$ to $t=0$.

\paragraph{Probability path and training objective.}
We use a linear Gaussian probability path, which corresponds to rectified flow matching~\cite{lipman2023flow}.
We draw Gaussian noise $\boldsymbol{\epsilon}\sim\mathcal{N}(\mathbf{0},\mathbf{I})$ and define a linear probability path
\begin{equation}
  \mathbf{z}_t=(1-t)\mathbf{z}_0+t\boldsymbol{\epsilon}, \quad t\in[0,1],
\end{equation}
where $t=1$ corresponds to pure noise and $t=0$ corresponds to the clean edge latent.
Our conditional model learns a time-dependent velocity field $\mathbf{v}_\theta(\mathbf{z}_t,t;\,x)$.
Although FLUX incorporates text conditioning, we omit explicit text-feature notation in our formulation because we use a fixed prompt throughout all experiments.
During training, we sample $t\sim\mathcal{U}(0,1)$ and minimize the objective
\begin{equation}
  \mathcal{L}_{\mathrm{FM}}=
  \mathbb{E}\Bigl[\bigl\|\mathbf{v}_\theta(\mathbf{z}_t,t;\,x)-(\boldsymbol{\epsilon}-\mathbf{z}_0)\bigr\|_2^2\Bigr].
\end{equation}

\paragraph{Sampling with discrete updates.}
At inference, we sample $\mathbf{z}_{t_0}\sim\mathcal{N}(\mathbf{0},\mathbf{I})$ at $t_0=1$ and integrate
$\frac{d\mathbf{z}_t}{dt}=\mathbf{v}_\theta(\mathbf{z}_t,t;\,x)$
from $t=1$ to $0$ with $K$ discrete steps.
Let $\{t_k\}_{k=0}^{K}$ be a decreasing schedule with $t_0=1$ and $t_K=0$.
A generic one-step update can be written as
\begin{equation}
  \mathbf{z}_{t_{k+1}}=\mathbf{z}_{t_k}+\Delta t_k\,\mathbf{v}_\theta(\mathbf{z}_{t_k},t_k;\,x),
  \qquad \Delta t_k=t_{k+1}-t_k<0,
\end{equation}
and the final prediction is obtained as $\hat{y}=D(\mathbf{z}_{t_K})$.
\subsection{Lightweight Adaptation via Condition Injection}
\label{sec:cond_inject}
We fine-tune an image-generation foundation model for edge detection by injecting image conditions using a small number of trainable parameters while keeping the backbone frozen.
Following EasyControl~\cite{zhang2025easycontrol}, we adapt the DiT-based foundation model by adding a plug-and-play Condition Injection LoRA that injects image conditions while keeping the original backbone weights frozen.
We encode and patchify the input image $x$ into condition tokens and inject them into each transformer block, optimizing only the LoRA parameters for edge detection.
A detailed description is provided in Appendix~\ref{app:condition_injection}.
%
\subsection{Edge-Specialized Training: Pixel-Space Objective and Lightweight Optimization}
\label{sec:pix_loss}
In edge detection, pixel misalignment hurts accuracy; prior fine-tuning~\cite{zhang2025easycontrol} ignores pixel errors.
To inject edge-specific supervision beyond $\mathcal{L}_{\mathrm{FM}}$, we add a pixel-space objective and efficient backpropagation.

Given $(\mathbf{z}_t,t)$, we form a one-step clean estimate
$\hat{\mathbf{z}}_0=\mathbf{z}_t-t\,\mathbf{v}_\theta(\mathbf{z}_t,t;\,x)$,
decode it as $\hat{y}=D(\hat{\mathbf{z}}_0)$, and apply an uncertainty-aware pixel-space loss.
Following \cite{ye2024diffusionedge}, we employ a weighted cross-entropy loss.
For each pixel $i$, the loss is defined as:
\begin{equation}
  \ell_i = \begin{cases}
    -\alpha \cdot \log(1 - \hat{y}_i), & \text{if } y_i = 0,        \\
    0,                                 & \text{if } 0 < y_i < \eta, \\
    -\beta \cdot \log \hat{y}_i,       & \text{otherwise},
  \end{cases}
\end{equation}
where $\hat{y}_i \in [0, 1]$ is the predicted edge probability at pixel $i$, $y_i$ is the ground truth edge probability (averaged over multiple annotations), and $\eta$ is a threshold to identify uncertain pixels.
The weights $\alpha$ and $\beta$ are set to balance the class imbalance:
\begin{equation}
  \alpha = \lambda \cdot \frac{|Y_+|}{|Y_+| + |Y_-|}, \quad
  \beta = \frac{|Y_-|}{|Y_+| + |Y_-|},
\end{equation}
where $|Y_+|$ and $|Y_-|$ denote the number of edge and non-edge pixels, respectively, and $\lambda$ is a hyperparameter.
We define the pixel-space loss as
$\mathcal{L}_{\mathrm{pix}}=\frac{1}{N_{\mathrm{valid}}}\sum_i \ell_i$,
where $N_{\mathrm{valid}}$ counts non-ambiguous pixels ($y_i=0$ or $y_i\ge\eta$). Our training objective is
\begin{equation}
  \mathcal{L}=\mathcal{L}_{\mathrm{FM}}+\sigma_t\,\mathcal{L}_{\mathrm{pix}}(\hat{y},y),
  \qquad \sigma_t=(1-t)^2.
\end{equation}

%

%
For efficiency and stability, we follow the uncertainty distillation idea in~\cite{ye2024diffusionedge}.
Thus, we use $D$ only in the forward pass to evaluate the pixel-space objective $\mathcal{L}_{\text{pix}}$.
Instead, we inject a proxy gradient $\mathbf g$ derived from the scalar loss $\mathcal{L}_{\text{pix}}$ and broadcastable to $\hat{\mathbf z}_0$, with $\mathbf g=\mathcal{L}_{\text{pix}}\,\mathbf 1$, using a custom autograd operator that overrides the backward gradient on $\hat{\mathbf z}_0$ as
\begin{equation}
  \frac{\partial \mathcal{L}_{\text{pix}}}{\partial \hat{\mathbf z}_0} \leftarrow \mathbf g,
  \qquad
  \nabla_\theta \mathcal{L}_{\text{pix}} \approx
  \frac{\partial \hat{\mathbf z}_0}{\partial \theta}^{\!\top}\mathbf g .
\end{equation}
This avoids the memory and compute overhead of decoder backpropagation while still allowing pixel-space supervision to influence the latent dynamics.
A detailed description is provided in Appendix~\ref{app:proxygrad}.

\subsection{Controllable Inference via Flow-Matching Guidance}
\label{sec:fm_guidance}
At inference, we exploit the composability of vector fields in Flow Matching and control conditioning strength via a guidance scale $\gamma$. Let $\mathbf{v}_{\mathrm{cond}}(\cdot;\,x)$ be the conditional dynamics from our adapted model, and let $\mathbf{v}_{\mathrm{base}}(\cdot;\,\varnothing)$ denote the unconditional dynamics provided by the frozen foundation model. Following \cite{phunyaphibarn2025unconditional}, we define the guided field as
\begin{equation}
  \mathbf{v}_\gamma=\mathbf{v}_{\mathrm{base}}(\cdot;\varnothing)+\gamma\Bigl(\mathbf{v}_{\mathrm{cond}}(\cdot;x)-\mathbf{v}_{\mathrm{base}}(\cdot;\varnothing)\Bigr),
  \label{eqn:cfg}
\end{equation}
and integrate the ODE using $\mathbf{v}_\gamma$.
We obtain $\mathbf{v}_{\mathrm{base}}$ from the original pre-trained FLUX.1-dev~\cite{fluxdev} which does not use the image condition $x$.
The fixed text prompt is applied in both $\mathbf{v}_{\mathrm{cond}}$ and $\mathbf{v}_{\mathrm{base}}$.
This design enables a single trained model to adjust the edge density via $\gamma$.

\section{Experiments}
\begin{table*}[t]
  \centering
  \small
  \setlength{\tabcolsep}{3.0pt}
  \renewcommand{\arraystretch}{1.0}

  \caption{Quantitative comparisons on BSDS500, NYUDv2, and BIPED.}
  \label{tab:edge_detection}

  \begin{tabular}{l|cccc|cccc|cccc}
    \hline
    \multirow{3}{*}{Methods}                 & \multicolumn{4}{c|}{BSDS500} & \multicolumn{4}{c|}{NYUDv2} & \multicolumn{4}{c}{BIPED}                                                                                                                                                                                            \\
    \cline{2-5} \cline{6-9} \cline{10-13}
                                             & \multicolumn{2}{c}{SEval}    & \multicolumn{2}{c|}{CEval}  & \multicolumn{2}{c}{SEval} & \multicolumn{2}{c|}{CEval} & \multicolumn{2}{c}{SEval} & \multicolumn{2}{c}{CEval}                                                                                                       \\
    \cline{2-3} \cline{4-5} \cline{6-7} \cline{8-9} \cline{10-11} \cline{12-13}
                                             & ODS                          & OIS                         & ODS                       & OIS                        & ODS                       & OIS                       & ODS            & OIS            & ODS            & OIS            & ODS            & OIS            \\
    \hline
    EDTER~\cite{pu2022edter}                 & 0.824                        & 0.841                       & 0.698                     & 0.706                      & 0.774                     & 0.789                     & 0.430          & 0.457          & 0.893          & 0.898          & --             & --             \\
    UAED~\cite{zhou2023uaed}                 & 0.829                        & 0.847                       & 0.722                     & 0.731                      & --                        & --                        & --             & --             & --             & --             & --             & --             \\
    EdgeSAM-S~\cite{yang2024edgesam}         & 0.838                        & 0.852                       & --                        & --                         & 0.783                     & 0.797                     & --             & --             & --             & --             & --             & --             \\
    EDMB~\cite{li2025edmb}                   & 0.837                        & 0.851                       & --                        & --                         & 0.774                     & 0.787                     & --             & --             & 0.903          & 0.907          & --             & --             \\
    DiffusionEdge~\cite{ye2024diffusionedge} & 0.834                        & 0.848                       & 0.749                     & 0.754                      & 0.761                     & 0.766                     & 0.732          & 0.738          & 0.899          & 0.901          & 0.889          & 0.897          \\
    GED~\cite{zhou2024ged}                   & \textbf{0.859}               & \textbf{0.875}              & $0.678^{*}$               & $0.692^{*}$                & $0.780^{*}$               & $0.797^{*}$               & $0.579^{*}$    & $0.609^{*}$    & $0.862^{*}$    & $0.885^{*}$    & $0.711^{*}$    & $0.743^{*}$    \\
    \hline
    Ours (K=1)                               & {0.857}                      & 0.869                       & 0.790                     & 0.797                      & 0.788                     & 0.797                     & 0.736          & 0.749          & 0.899          & 0.906          & 0.858          & 0.869          \\
    Ours (K=5)                               & {0.857}                      & 0.873                       & \textbf{0.807}            & \textbf{0.819}             & \textbf{0.791}            & 0.799                     & {0.773}        & \textbf{0.784} & {0.903}        & 0.909          & \textbf{0.892} & \textbf{0.902} \\
    Ours (K=50)                              & 0.855                        & 0.873                       & 0.805                     & \textbf{0.819}             & 0.780                     & 0.788                     & 0.770          & 0.777          & 0.900          & 0.906          & 0.889          & 0.897          \\
    Ours (K=50, w/ CFG)                      & {0.857}                      & {0.874}                     & 0.803                     & 0.817                      & \textbf{0.791}            & \textbf{0.800}            & \textbf{0.774} & 0.780          & \textbf{0.908} & \textbf{0.913} & 0.887          & 0.893          \\
    \hline
  \end{tabular}
  \vspace{2pt}

  \small{$^{*}$Our own implementation of GED, as no official code was released.}
\end{table*}
\begin{table}[t]
  \centering
  \fontsize{9}{10.8}\selectfont
  \setlength{\tabcolsep}{3.0pt}
  \renewcommand{\arraystretch}{1.0}
  \caption{
    Quantitative comparison on the BIPED dataset.
    The column headers denote the percentage of training samples.
  }
  \label{tab:biped_fewshot_result}
  \begin{tabular}{l|cccc}
    \toprule
    \multirow{2}{*}{Method}
                   & \multicolumn{4}{c}{10\% (20 images)}                                                    \\
    \cline{2-5}
                   & \multicolumn{2}{c}{SEval}
                   & \multicolumn{2}{c}{CEval}                                                               \\
    \cline{2-5}
                   & ODS                                  & OIS            & ODS            & OIS            \\
    \midrule
    DiffusionEdge  & 0.853                                & 0.854          & 0.851          & 0.857          \\
    GED            & 0.862                                & 0.874          & 0.746          & 0.771          \\
    \hline
    Ours ($K=5$)   & 0.878                                & 0.883          & \textbf{0.875} & \textbf{0.883} \\
    Ours           & 0.877                                & 0.883          & \textbf{0.875} & \textbf{0.883} \\
    Ours  (w/ CFG) & \textbf{0.881}                       & \textbf{0.888} & 0.852          & 0.856          \\
    \bottomrule
  \end{tabular}
\end{table}
\label{sec:experiments}
\subsection{Experimental Setup}
\paragraph{Implementation details.}
We instantiate a model by adapting FLUX.1-dev~\cite{batifol2025flux, fluxdev} to edge generation using EasyControl~\cite{zhang2025easycontrol} as the condition-injection, and extending it with our edge-specialized objectives.
We keep the backbone frozen and train only the lightweight condition-injection parameters.
The number of inference steps is $K=50$ unless otherwise noted.
Detailed descriptions of training settings, data preprocessing and inference settings are provided in Appendix~\ref{app:exp_setup}.

\paragraph{Comparison Methods.}
We mainly compare against DiffusionEdge~\cite{ye2024diffusionedge} and GED~\cite{zhou2024ged}.
DiffusionEdge generates edge maps via iterative diffusion-based refinement.
GED adapts an image-generation foundation model to edge detection and predicts edge maps in one step.
GED originally uses image-specific text prompts.
For fairness, we use a single fixed text prompt for GED and our method across each dataset.
We also include recent edge detection methods~\cite{pu2022edter, zhou2023uaed, yang2024edgesam, li2025edmb} as supplementary baselines.


\paragraph{Datasets.}
To assess cross-domain performance, we evaluate on standard edge detection and wall-boundary detection, the latter being a domain-specific instance of edge detection in architectural floor plans.
We use BSDS500~\cite{MartinFTM01}, NYUDv2~\cite{Silberman:ECCV12}, and BIPED~\cite{soria2020dexined} for general edge detection, and CubiCasa~\cite{kalervo2019cubicasa5k} for wall-boundary detection.
CubiCasa contains 5,000 samples with various kinds of annotations; we use only the wall polygon labels.
Additional dataset details are provided in Appendix~\ref{app:dataset}.

\paragraph{Evaluation Metrics.}
For BSDS500, NYUDv2, and BIPED, we evaluate edge detection with precision ($P$), recall ($R$), and F-score ($F$).
We threshold the predicted edge probability map to obtain a binary map, and report \textbf{ODS} (a single global threshold for the dataset) and \textbf{OIS} (the best threshold per image).
The F-score is $F=\frac{2PR}{P+R}$. Matching tolerances are 0.011 for NYUDv2 and 0.0075 for the others.
Following prior work~\cite{ye2024diffusionedge}, we report Standard evaluation (\textbf{SEval}) and Crispness evaluation (\textbf{CEval}).
SEval applies standard post-processing (NMS + morphological thinning), whereas CEval is computed on raw predictions.
For CubiCasa, we report the boundary F-measure and wall-region IoU, both measured on raw predictions without post-processing.

\subsection{Results on Standard Edge Detection}
\begin{figure}[t]
  \centering
  \includegraphics[width=0.99\linewidth]{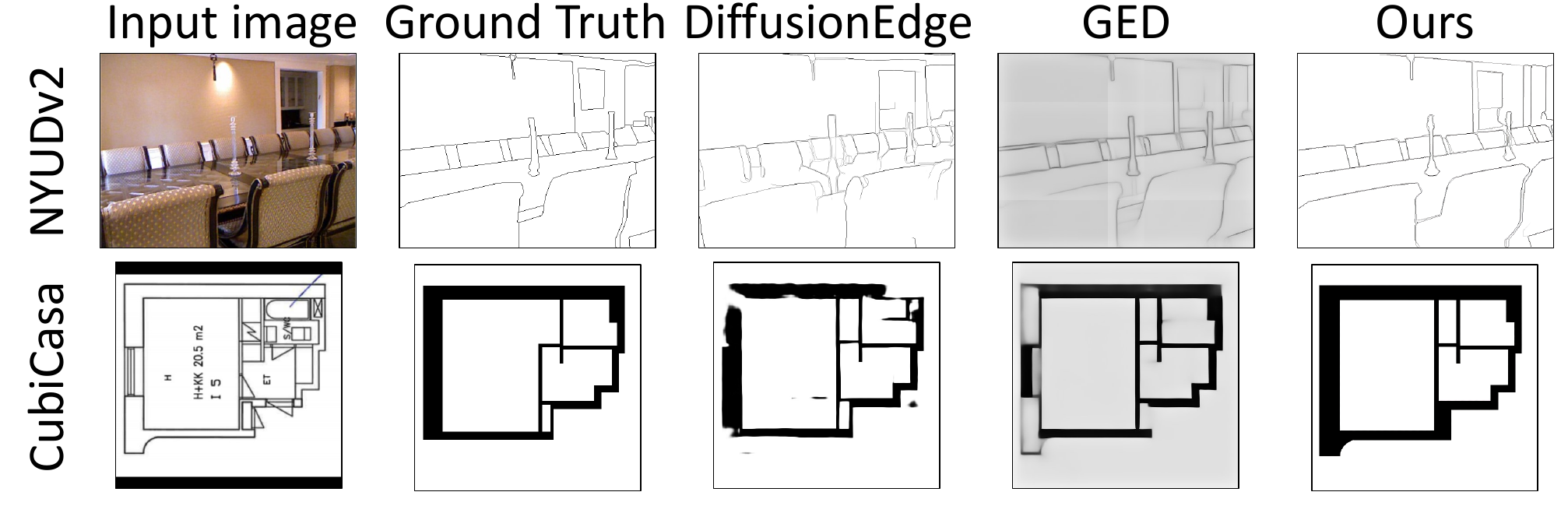}
  \caption{
    Qualitative results.
    Top row shows the results on NYUDv2 and bottom row shows the results on CubiCasa.}
  \label{fig:biped}
\end{figure}

Tab.~\ref{tab:edge_detection} summarizes the results of standard edge detection.
Our method achieves competitive or better performance.
It consistently outperforms DiffusionEdge under both SEval and CEval, suggesting foundation-model priors yield accurate and crisp edges.
While GED is competitive under SEval on some datasets, our method outperforms it by a wide margin under CEval; multi-step generation sharpens raw predictions.
The top row of qualitative results (Fig.~\ref{fig:biped}) further show crisper and more accurate edges than prior methods, which often produce thicker/blurrier edges or miss fine details.
Additional qualitative results are provided in Appendix~\ref{app:additional_qualitative_results}.

These results also highlight the benefit of multi-step generation in crisp edge detection.
While $K=1$ and $K>1$ yield similar SEval scores, $K>1$ substantially outperforms under CEval.
Fig.~\ref{fig:stepnum} demonstrates this effect qualitatively, with additional examples provided in Appendix~\ref{app:inference_steps}.

To assess data efficiency, we fine-tune on BIPED using only 10\% of the training set.
Tab.~\ref{tab:biped_fewshot_result} shows that our method substantially outperforms prior methods, demonstrating the benefit of foundation-model priors with limited training data.
\begin{figure}[t]
  \centering
  \includegraphics[width=0.85\linewidth]{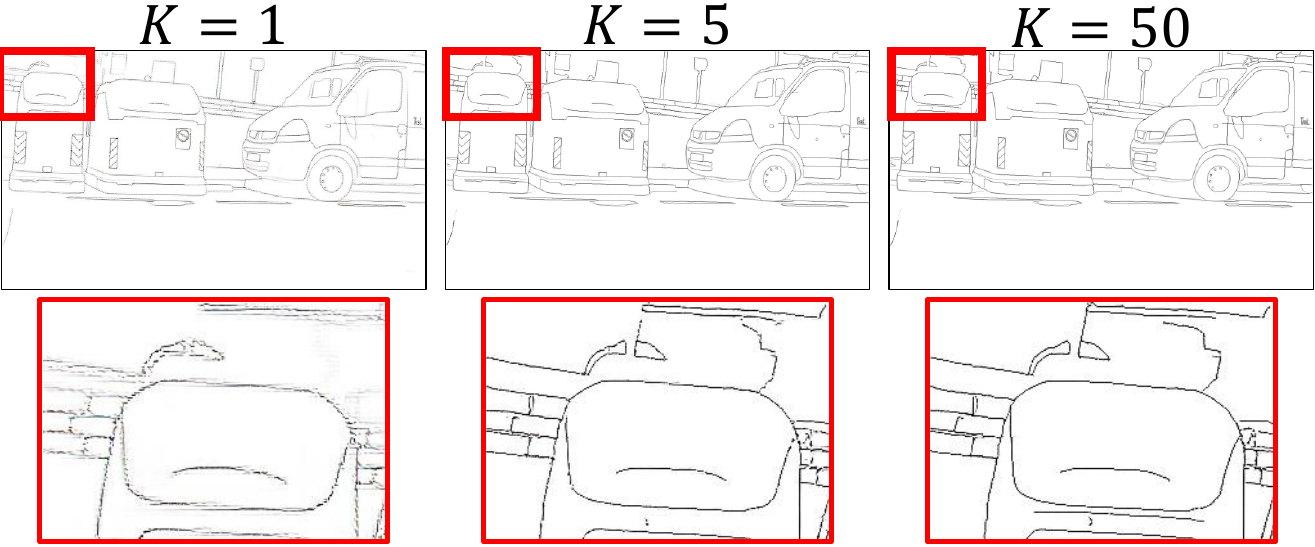}
  \caption{
    Qualitative comparison of different inference steps $K$ on BIPED.
    Increasing $K$ sharpens edges and recovers fine details.
  }
  \label{fig:stepnum}
\end{figure}
\subsection{Results on Wall-Boundary Detection.}
Tab.~\ref{tab:cubi_result} reports results of wall-boundary detection.
Our method is competitive with, or outperforms, prior methods in both IoU and F-score.
Notably, our method maintains strong performance even with 1\% or 10\% of the training data (42 and 420 images, respectively).
The bottom row of Fig.~\ref{fig:biped} further shows that our method directly generates vectorization-ready edge maps without post-processing.
\begin{table}[t]
  \centering
  \small
  \setlength{\tabcolsep}{3.0pt}
  \renewcommand{\arraystretch}{1.0}
  \caption{
    Quantitative comparison of wall detection on the CubiCasa dataset (4,200 training images).
    The column headers denote the percentage of training samples used to train the models.
  }
  \label{tab:cubi_result}
  \begin{tabular}{l|cc|cc|cc}
    \toprule
    \multirow{2}{*}{Method} & \multicolumn{2}{c|}{100\%} & \multicolumn{2}{c|}{10\%} & \multicolumn{2}{c}{1\%}                                                    \\
    \cline{2-7}
                            & IoU                        & F-Score                   & IoU                     & F-Score        & IoU            & F-Score        \\
    \midrule
    DiffusionEdge           & {0.793}                    & {0.881}                   & 0.757                   & 0.858          & 0.642          & 0.775          \\
    GED                     & 0.771                      & 0.867                     & 0.738                   & 0.846          & 0.699          & 0.819          \\
    \hline
    Ours                    & 0.760                      & {0.860}                   & {0.767}                 & {0.864}        & {0.739}        & {0.846}        \\
    Ours (w/ CFG)           & \textbf{0.794}             & \textbf{0.882}            & \textbf{0.780}          & \textbf{0.873} & \textbf{0.742} & \textbf{0.848} \\
    \bottomrule
  \end{tabular}
\end{table}
\subsection{Ablation Study.}
\begin{wrapfigure}{r}{0.485\linewidth}
  \centering
  \includegraphics[width=\linewidth]{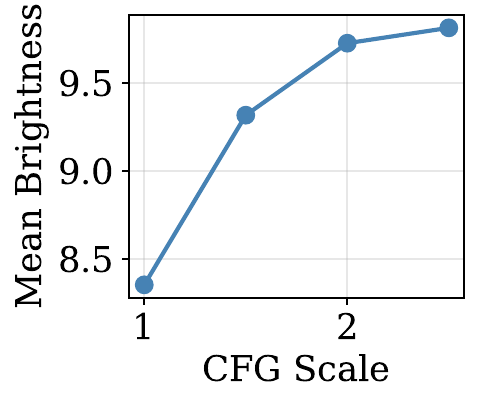}
  \caption{Mean Brightness vs $\gamma$.}
  \label{fig:curve}
\end{wrapfigure}

\begin{figure}[t]
  \centering
  \includegraphics[width=0.92\linewidth]{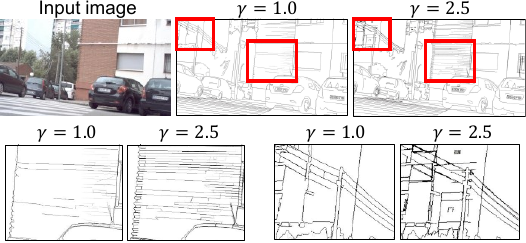}
  \caption{Qualitative effect of guidance scale $\gamma$ on BIPED.
  }
  \label{fig:guidance_qual}
\end{figure}

\begin{figure}[t]
  \centering
  \includegraphics[width=0.6\linewidth]{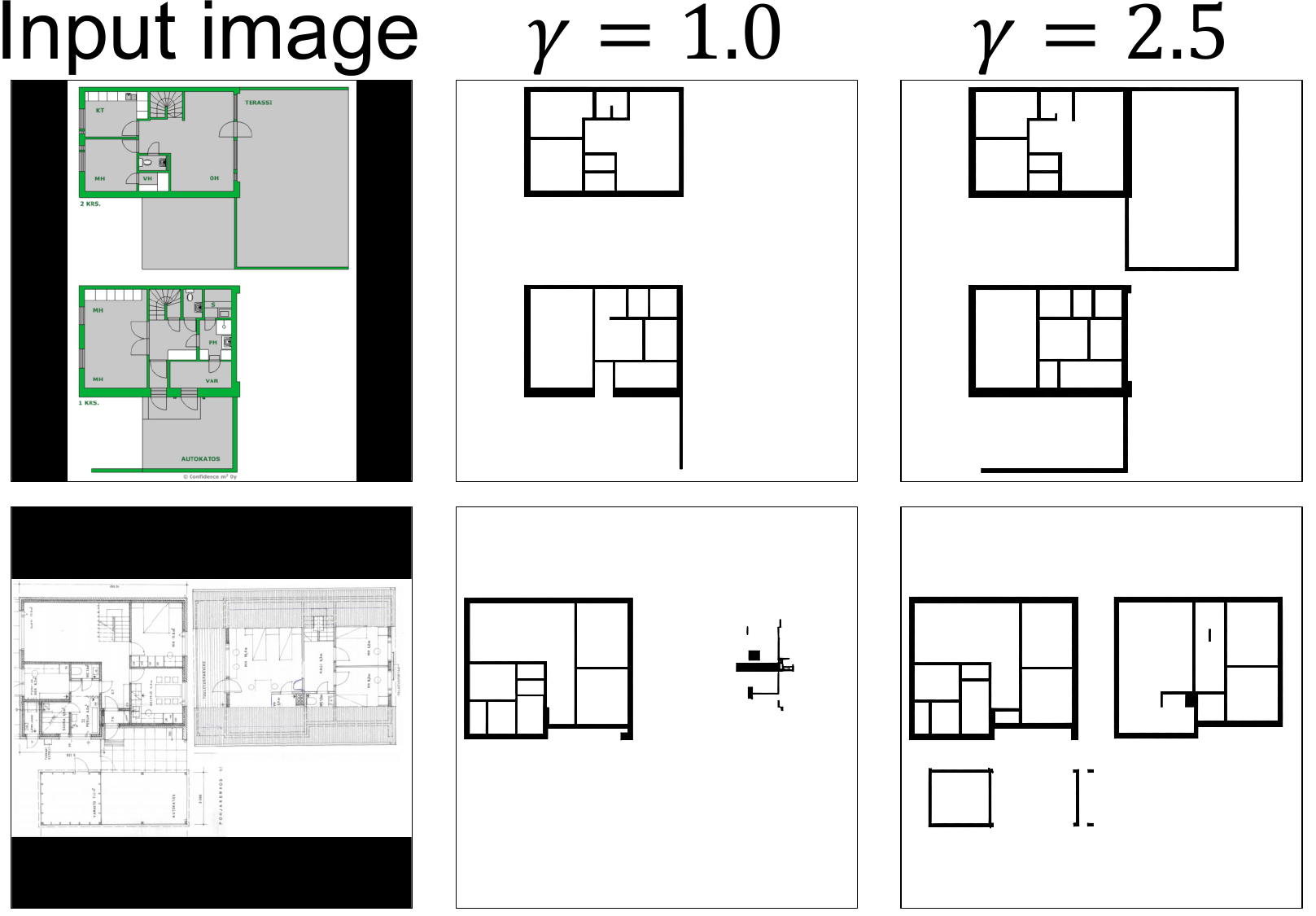}
  \caption{Qualitative effect of guidance scale $\gamma$ on CubiCasa.
  }
  \label{fig:guidance_qual_cubi}
\end{figure}

\paragraph{Effectiveness of a guidance scale.}
We study the effectiveness of the guidance scale $\gamma$.
Fig.~\ref{fig:curve} shows the mean brightness of predicted edge maps on the BIPED dataset as $\gamma$ varies; it indicates that larger $\gamma$ monotonically increases mean brightness.
This trend indicates that stronger guidance produces denser edge responses, i.e., a larger fraction of pixels are classified as edges.
%
%
This trend is further supported by the qualitative results in Figs.~\ref{fig:guidance_qual}, \ref{fig:guidance_qual_cubi}. With a larger
$\gamma$, the model produces denser and finer-grained edges, recovering subtle structures and thin contours.
In contrast, reducing $\gamma$ suppresses these small details and yields cleaner maps that retain only the most salient boundaries.
Importantly, the best operating point depends on the target application, and our results demonstrate that a single trained model can select this operating point simply by adjusting $\gamma$.
%
%

\paragraph{Effectiveness of a pixel loss.}
We study the effect of the pixel-space loss $\mathcal{L}_{\mathrm{pix}}$ by removing it while keeping the rest unchanged.
\begin{wraptable}{r}{0.57\linewidth}
  \centering
  \caption{The effectiveness of $\mathcal{L}_{\mathrm{pix}}$.}
  \label{tab:ablation_pix}
  \scriptsize
  \setlength{\tabcolsep}{2pt}
  \renewcommand{\arraystretch}{1.0}
  \resizebox{\linewidth}{!}{%
    \begin{tabular}{@{}c|cc|cc@{}}
      \hline
                                       & \multicolumn{2}{c|}{BIPED} & \multicolumn{2}{c}{CubiCasa}                                   \\
      \cline{2-5}
                                       & \multicolumn{2}{c|}{CEval} & \multicolumn{2}{c}{}                                           \\
                                       & ODS                        & OIS                          & IoU            & F-score        \\
      \hline
      w/o $\mathcal{L}_{\mathrm{pix}}$ & 0.879                      & 0.881                        & 0.785          & 0.877          \\
      w $\mathcal{L}_{\mathrm{pix}}$   & \textbf{0.887}             & \textbf{0.893}               & \textbf{0.794} & \textbf{0.882} \\
      \hline
    \end{tabular}%
  }
\end{wraptable}
As shown in Tab.~\ref{tab:ablation_pix}, disabling $\mathcal{L}_{\mathrm{pix}}$ consistently degrades performance across datasets and the metrics, demonstrating the benefit of edge-specialized training objectives.
\section{Discussion}
\label{sec:discussion}
A limitation of this study is that our evaluation is restricted to FLUX.1-dev~\cite{batifol2025flux, fluxdev} with EasyControl~\cite{zhang2025easycontrol}; we do not evaluate other backbones or adaptation framework.
Nevertheless, our key insight is that we connect iterative generative priors and inference-time controllability in foundation models to edge-detection objectives.
As next-generation backbones achieve higher-fidelity generation, simply swapping in such models may improve edge correctness, crispness, and data efficiency.
Given the rapid advances in image-generation foundation models~\cite{esser2024scaling,Qin2025ICCV,ICLR2025ACE} and parameter-efficient adaptation methods~\cite{Mao2025ICCV}, we expect our framework to benefit from these improvements.

At inference time, our method uses guidance as in Eq.~(\ref{eqn:cfg}), and adjusts output characteristics by varying the guidance scale $\gamma$.
Importantly, this controllability is likely to be preserved as long as generation relies on a conditional, multi-step iterative sampling procedure, even as the backbone architecture evolves.
As a result, our approach can benefit from improvements in both model quality and efficiency.
%
\section{Conclusion}
\label{sec:conclusion}
We present EasyControlEdge to specialize foundation models for crisp, data efficient, and controllable edge detection.
Our method directly generates crisp, well-localized boundaries and offers inference-time control of edge density through guidance scaling; it also demonstrates strong data efficiency, particularly in experiments with limited training data and when evaluated without any post-processing.
%
\bibliographystyle{IEEEbib}
{\small
  \bibliography{strings,refs}

@inproceedings{kalervo2019cubicasa5k,
  title        = {Cubicasa5k: A dataset and an improved multi-task model for floorplan image analysis},
  author       = {Kalervo, Ahti and others},
  booktitle    = {Scandinavian Conference on Image Analysis},
  year         = {2019},
  organization = {Springer}
}

@inproceedings{MartinFTM01,
  author    = {D. Martin and C. Fowlkes and D. Tal and J. Malik},
  title     = {A Database of Human Segmented Natural Images and its
               Application to Evaluating Segmentation Algorithms and
               Measuring Ecological Statistics},
  booktitle = {ICCV},
  year      = {2001}
}

@inproceedings{Silberman:ECCV12,
  author    = {Silberman, Nathan and Hoiem, Derek and Kohli, Pushmeet and Fergus, Rob},
  title     = {Indoor Segmentation and Support Inference from RGBD Images},
  booktitle = {ECCV},
  year      = {2012}
}

@inproceedings{soria2020dexined,
  title     = {Dense Extreme Inception Network: Towards a Robust CNN Model for Edge Detection},
  author    = {Xavier Soria and Edgar Riba and Angel Sappa},
  booktitle = {WACV},
  year      = {2020}
}

@inproceedings{ye2024diffusionedge,
  title     = {Diffusionedge: Diffusion probabilistic model for crisp edge detection},
  author    = {Ye, Yunfan and Xu, Kai and Huang, Yuhang and Yi, Renjiao and Cai, Zhiping},
  booktitle = {AAAI},
  year      = {2024}
}

@article{canny1986,
  title   = {A Computational Approach to Edge Detection},
  author  = {Canny, John},
  journal = {TPAMI},
  year    = {1986}
}

@inproceedings{pu2022edter,
  title     = {EDTER: Edge Detection with Transformer},
  author    = {Pu, Mengyang and Huang, Yaping and Guan, Qingji and Huang, Li},
  booktitle = {CVPR},
  year      = {2022}
}

@inproceedings{zhou2023uaed,
  title     = {The Treasure Beneath Multiple Annotations: An Uncertainty-Aware Edge Detector},
  author    = {Zhou, Caixia and Huang, Yaping and Pu, Mengyang and Guan, Qingji and Huang, Li and Ling, Haibin},
  booktitle = {CVPR},
  year      = {2023}
}

@article{yang2024edgesam,
  title   = {Boosting Deep Unsupervised Edge Detection via Segment Anything Model},
  author  = {Yang, Wenya and Pan, Junju and Yu, Haibin and Li, Qing and Ling, Haibin},
  journal = {IEEE Transactions on Industrial Informatics},
  year    = {2024}
}

@misc{zhou2024ged,
  title        = {GED: Generative Edge Detection with Diffusion Models},
  author       = {Zhou, Caixia and Huang, Yaping and Pu, Mengyang and Guan, Qingji and Deng, Ruoxi and Ling, Haibin},
  howpublished = {arXiv preprint},
  year         = {2024}
}

@inproceedings{lipman2023flow,
  title     = {Flow Matching for Generative Modeling},
  author    = {Yaron Lipman and Ricky T. Q. Chen and Heli Ben-Hamu and Maximilian Nickel and Matthew Le},
  booktitle = {ICLR},
  year      = {2023}
}

@article{batifol2025flux,
  title   = {FLUX. 1 Kontext: Flow Matching for In-Context Image Generation and Editing in Latent Space},
  author  = {Batifol, Stephen and Blattmann, Andreas and Boesel, Frederic and Consul, Saksham and Diagne, Cyril and Dockhorn, Tim and English, Jack and English, Zion and Esser, Patrick and Kulal, Sumith and others},
  journal = {arXiv preprint},
  year    = {2025}
}

@inproceedings{rombach2022ldm,
  title     = {High-Resolution Image Synthesis with Latent Diffusion Models},
  author    = {Rombach, Robin and Blattmann, Andreas and Lorenz, Dominik and Esser, Patrick and Ommer, Bj{\"o}rn},
  booktitle = {CVPR},
  year      = {2022}
}

@inproceedings{ho2021classifier,
  title     = {Classifier-Free Diffusion Guidance},
  author    = {Ho, Jonathan and Salimans, Tim},
  booktitle = {NeurIPS Workshop},
  year      = {2021}
}

@inproceedings{zhang2025easycontrol,
  title     = {Easycontrol: Adding efficient and flexible control for diffusion transformer},
  author    = {Zhang, Yuxuan and Yuan, Yirui and Song, Yiren and Wang, Haofan and Liu, Jiaming},
  booktitle = {ICCV},
  year      = {2025}
}

@inproceedings{li2025edmb,
  title     = {EDMB: Edge Detector with Mamba},
  author    = {Li, Yachuan and Poma, Xavier Soria and Bai, Yun and Xiao, Qian and Yang, Chaozhi and Li, Guanlin and Li, Zongmin},
  booktitle = {WACV},
  year      = {2025}
}

@article{phunyaphibarn2025unconditional,
  title   = {Unconditional Priors Matter! Improving Conditional Generation of Fine-Tuned Diffusion Models},
  author  = {Phunyaphibarn, Prin and Lee, Phillip Y and Kim, Jaihoon and Sung, Minhyuk},
  journal = {arXiv preprint},
  year    = {2025}
}

@inproceedings{cheng2020boundary,
  title     = {Boundary-preserving mask r-cnn},
  author    = {Cheng, Tianheng and Wang, Xinggang and Huang, Lichao and Liu, Wenyu},
  booktitle = {ECCV},
  year      = {2020}
}

@inproceedings{wang2017deepcrisp,
  author    = {Yupei Wang and Xin Zhao and Kaiqi Huang},
  title     = {Deep Crisp Boundaries},
  booktitle = {CVPR},
  year      = {2017}
}

@article{zhou2024notricks,
  author  = {Jianhang Zhou and Hongwei Zhao and Haoyu Zhao and Pengyu Mu and Long Xing and Mingsi Sun},
  title   = {No tricks no bluff, focusing on localizing crisp boundaries in image media},
  journal = {Neurocomputing},
  year    = {2024}
}

@article{huan2021unmixing,
  title   = {Unmixing convolutional features for crisp edge detection},
  author  = {Huan, Linxi and Xue, Nan and Zheng, Xianwei and He, Wei and Gong, Jianya and Xia, Gui-Song},
  journal = {TPAMI},
  year    = {2021}
}

@article{ye2023delving,
  title   = {Delving into crispness: Guided label refinement for crisp edge detection},
  author  = {Ye, Yunfan and Yi, Renjiao and Gao, Zhirui and Cai, Zhiping and Xu, Kai},
  journal = {TIP},
  year    = {2023}
}

@inproceedings{esser2024scaling,
  title     = {Scaling rectified flow transformers for high-resolution image synthesis},
  author    = {Esser, Patrick and Kulal, Sumith and Blattmann, Andreas and Entezari, Rahim and M{\"u}ller, Jonas and Saini, Harry and Levi, Yam and Lorenz, Dominik and Sauer, Axel and Boesel, Frederic and others},
  booktitle = {ICML},
  year      = {2024}
}

@inproceedings{xu2025ootdiffusion,
  title     = {Ootdiffusion: Outfitting fusion based latent diffusion for controllable virtual try-on},
  author    = {Xu, Yuhao and Gu, Tao and Chen, Weifeng and Chen, Arlene},
  booktitle = {AAAI},
  year      = {2025}
}

@article{hu2022lora,
  title   = {Lora: Low-rank adaptation of large language models.},
  author  = {Hu, Edward J and Shen, Yelong and Wallis, Phillip and Allen-Zhu, Zeyuan and Li, Yuanzhi and Wang, Shean and Wang, Lu and Chen, Weizhu and others},
  journal = {ICLR},
  year    = {2022}
}

@inproceedings{qiu2025adaptively,
  title     = {Adaptively Distilled ControlNet: Accelerated Training and Superior Sampling for Medical Image Synthesis},
  author    = {Qiu, Kunpeng and Zhou, Zhiying and Guo, Yongxin},
  booktitle = {MICCAI},
  year      = {2025}
}

@inproceedings{Mao2025ICCV,
  author    = {Mao, Chaojie and Zhang, Jingfeng and Pan, Yulin and Jiang, Zeyinzi and Han, Zhen and Liu, Yu and Zhou, Jingren},
  title     = {ACE++: Instruction-Based Image Creation and Editing via Context-Aware Content Filling},
  booktitle = {ICCVW},
  year      = {2025}
}

@inproceedings{Qin2025ICCV,
  author    = {Qin, Qi and Zhuo, Le and Xin, Yi and Du, Ruoyi and Li, Zhen and Fu, Bin and Lu, Yiting and Li, Xinyue and Liu, Dongyang and Zhu, Xiangyang and Beddow, Will and Millon, Erwann and Perez, Victor and Wang, Wenhai and Qiao, Yu and Zhang, Bo and Liu, Xiaohong and Li, Hongsheng and Xu, Chang and Gao, Peng},
  title     = {Lumina-Image 2.0: A Unified and Efficient Image Generative Framework},
  booktitle = {ICCV},
  year      = {2025}
}

@inproceedings{ICLR2025ACE,
  title     = {ACE: All-round Creator and Editor Following Instructions via Diffusion Transformer},
  author    = {Han, Zhen and Jiang, Zeyinzi and Pan, Yulin and Zhang, Jingfeng and Mao, Chaojie and Xie, Chen-Wei and Liu, Yu and Zhou, Jingren},
  booktitle = {ICLR},
  year      = {2025}
}

@article{huang2025comprehensive,
  title     = {Comprehensive review of edge and contour detection: From traditional methods to recent advances},
  author    = {Huang, Qinyuan and Huang, Jiaxiong},
  journal   = {Neural Computing and Applications},
  year      = {2025},
  publisher = {Springer}
}

@misc{fluxdev,
  author       = {Black Forest Labs},
  title        = {FLUX.1-dev},
  year         = {2025},
  howpublished = {\url{https://huggingface.co/black-forest-labs/FLUX.1-dev}}
}

@inproceedings{liu2017raster,
  title     = {Raster-to-vector: Revisiting floorplan transformation},
  author    = {Liu, Chen and Wu, Jiajun and Kohli, Pushmeet and Furukawa, Yasutaka},
  booktitle = {ICCV},
  year      = {2017}
}

@inproceedings{zeng2019deep,
  title     = {Deep floor plan recognition using a multi-task network with room-boundary-guided attention},
  author    = {Zeng, Zhiliang and Li, Xianzhi and Yu, Ying Kin and Fu, Chi-Wing},
  booktitle = {ICCV},
  year      = {2019}
}

@article{jing2022recent,
  title   = {Recent advances on image edge detection: A comprehensive review},
  author  = {Jing, Junfeng and Liu, Shenjuan and Wang, Gang and Zhang, Weichuan and Sun, Changming},
  journal = {Neurocomputing},
  year    = {2022}
}

@inproceedings{loshchilovdecoupled,
  title     = {Decoupled Weight Decay Regularization},
  author    = {Loshchilov, Ilya and Hutter, Frank},
  booktitle = {International Conference on Learning Representations},
  year      = {2019}
}
}

\clearpage
\appendix

\section{Experimental Setup and Dataset Details}
\label{app:exp_setup_dataset}
\subsection{Experimental Setup}
\label{app:exp_setup}
\paragraph{Training Settings.}
All models are trained on an NVIDIA H200 GPU using AdamW~\cite{loshchilovdecoupled} with a learning rate of $10^{-4}$.
We train for 100,000 iterations with a batch size of $4$.
We set $\eta$ and $\lambda$ to $0.3$ and $1.1$.

\paragraph{Data Preprocessing.}
During training, input images are processed to a fixed spatial resolution of $512 \times 512$.
Specifically, if an image dimension is less than $512$, we pad the image such that the shorter side becomes $512$.
Subsequently, we apply a random crop of size $512 \times 512$.
For data augmentation, we apply only random horizontal flipping.
Specifically for the CubiCasa dataset~\cite{kalervo2019cubicasa5k}, we additionally employ random vertical flipping.

\paragraph{Inference.}
Unless otherwise specified, we run sampling for $K=50$ steps at inference time.
We apply CFG~\cite{ho2021classifier, phunyaphibarn2025unconditional}, using a guidance scale $\gamma$ of 2.0 for edge detection and 2.5 for wall-boundary detection.
For CFG, we treat an FLUX.1-dev~\cite{fluxdev} without any additional fine-tuning as $\mathbf{v}_{\mathrm{base}}$.
We keep the prompt fixed for all inputs.
For CubiCasa dataset, we use:
\begin{lstlisting}[basicstyle=\ttfamily\small, breaklines=true, frame=single]
Orthographic top-down residential floor plan, walls-only skeleton map. White solid walls on black background, thick outer walls, thin interior partitions. Clean CAD-like vector look.
\end{lstlisting}
For other datasets, we use:
\begin{lstlisting}[basicstyle=\ttfamily\small, breaklines=true, frame=single]
A line-art style image of a real-world scene, featuring clear object outlines with a pure black background and crisp white edges, no shading, and realistic proportions.
\end{lstlisting}
\begin{figure}[t]
  \centering
  \includegraphics[width=\columnwidth]{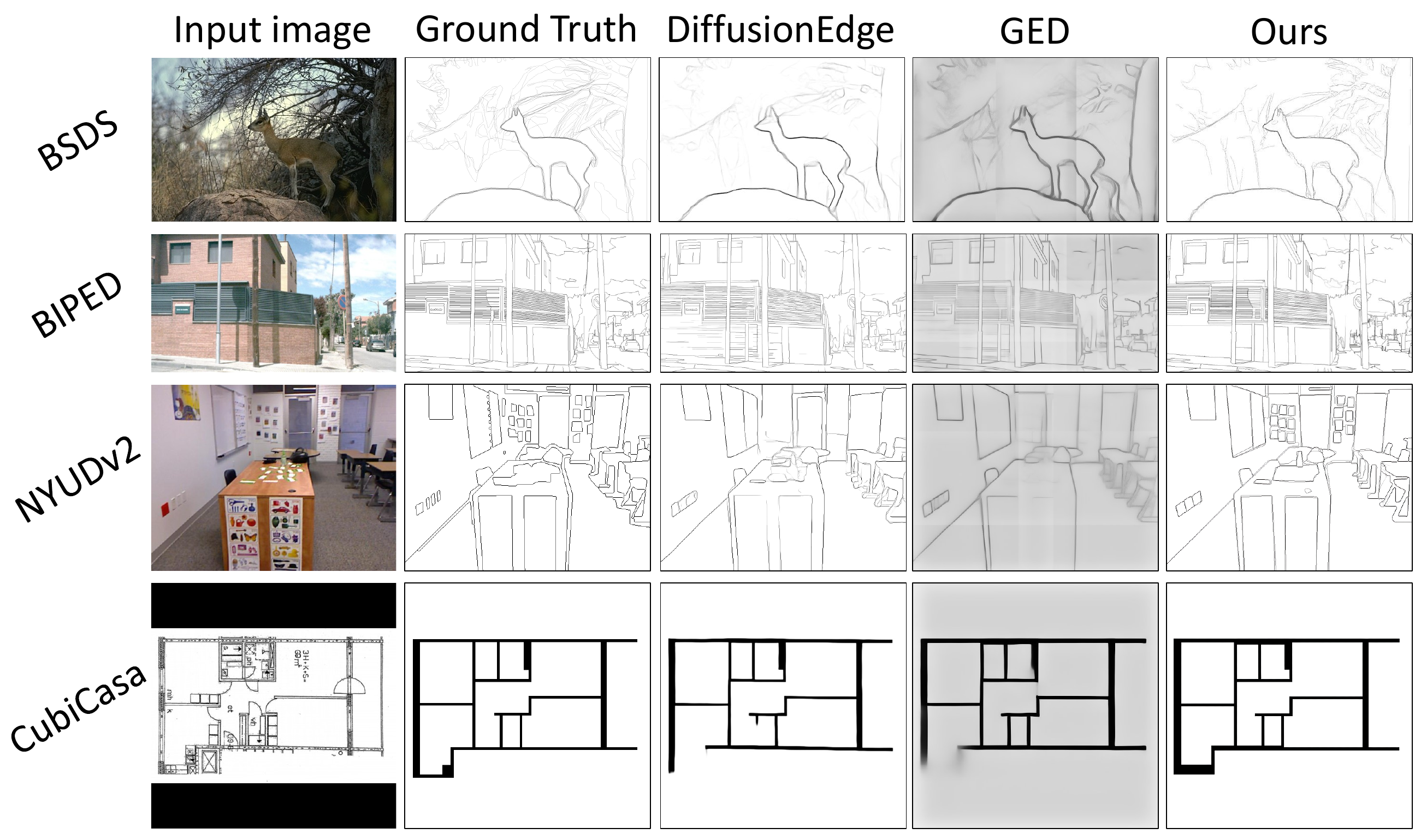}
  \caption{
    Additional qualitative results on BSDS500, NYUDv2, BIPED, and CubiCasa datasets.
    Our method consistently produces crisp edge maps across diverse scene types.}
  \label{fig:appx_qual}
\end{figure}

\subsection{Datasets}
\label{app:dataset}
\paragraph{BSDS500~\cite{MartinFTM01}.}
BSDS500 contains natural images with multiple human-annotated boundaries. The ground truth is computed by taking the average of these annotations. It consists of 200, 100, and 200 images for the training, validation, and test sets, respectively.

\paragraph{NYUDv2~\cite{Silberman:ECCV12}.}
NYUDv2 is comprised of indoor scenes accompanied by RGB-D data. The dataset is divided into 381 training, 414 validation, and 654 testing images.

\paragraph{BIPED~\cite{soria2020dexined}.}
BIPED features high-resolution images with crisp, fine-grained edge annotations. It contains a training set of 200 images and a testing set of 50 images.

\paragraph{CubiCasa~\cite{kalervo2019cubicasa5k}.}
CubiCasa provides 5,000 samples with versatile ground truth labels covering various architectural elements.
Since our focus is on wall boundaries, we exclusively use the wall polygon annotations.
For data preprocessing, we resize all images such that the longest dimension is 512 pixels while maintaining the aspect ratio. We then pad the shorter dimension to achieve a final fixed square resolution of $512 \times 512$.

\section{Details of our method.}
\subsection{Details of Lightweight Adaptation via Condition Injection}
\label{app:condition_injection}
This appendix formalizes the lightweight adaptation via condition injection mentioned in the main text.
Following EasyControl~\cite{zhang2025easycontrol}, we adapt the DiT-based foundation model~\cite{fluxdev} by adding a plug-and-play Condition Injection LoRA~\cite{hu2022lora} that injects image conditions while keeping the original backbone weights frozen.
We encode and patchify the input image $x$ into condition tokens and inject them into each transformer block, optimizing only the LoRA parameters for edge detection.

Formally, we view each transformer block as operating on three groups of tokens:
text tokens $Z_t$, diffusion/noise tokens $Z_n$, and condition tokens $Z_c$ (obtained by encoding and patchifying $x$).
In standard self-attention, each group is projected into queries, keys, and values by shared projection matrices:
{\fontsize{9.5}{11.4}\selectfont
\begin{equation}
  Q_i = W_Q Z_i,\quad K_i = W_K Z_i,\quad V_i = W_V Z_i,\qquad i\in\{t,n,c\}.
  \label{eq:qkv_shared}
\end{equation}
}%
To selectively enhance only the condition pathway, we introduce a low-rank residual (LoRA)~\cite{hu2022lora} on the condition branch projections:
{\fontsize{9.5}{11.4}\selectfont
\begin{equation}
  \Delta Q_c = B_Q A_Q Z_c,\quad
  \Delta K_c = B_K A_K Z_c,\quad
  \Delta V_c = B_V A_V Z_c,
  \label{eq:lora_delta}
\end{equation}
}%
where $A_{\{\cdot\}},B_{\{\cdot\}}\in\mathbb{R}^{d\times r}$ are trainable low-rank factors with $r\ll d$.
The condition-branch features are then updated as
\begin{equation}
  Q'_c = Q_c + \Delta Q_c,\quad
  K'_c = K_c + \Delta K_c,\quad
  V'_c = V_c + \Delta V_c,
  \label{eq:qkv_cond_updated}
\end{equation}
whereas the text and noise branches remain unchanged:
\begin{equation}
  Q'_i = Q_i,\quad K'_i = K_i,\quad V'_i = V_i,\qquad i\in\{t,n\}.
  \label{eq:qkv_frozen}
\end{equation}

Equations~\eqref{eq:lora_delta}--\eqref{eq:qkv_frozen} imply that we freeze the backbone projections $\{W_Q,W_K,W_V\}$ and learn only low-rank updates on the condition branch.
This selectively adjusts how condition tokens interact with other tokens through attention (via $Q'_c,K'_c$) and what information they contribute (via $V'_c$), while preserving the original text/noise representations.
Consequently, we obtain a parameter-efficient, plug-and-play conditioning mechanism suitable for edge-guided adaptation.

\subsection{Decoder-free proxy-gradient injection}
\label{app:proxygrad}

This section details the implementation of the decoder-free proxy-gradient injection~\cite{ye2024diffusionedge} used in Sec.~2.3.
Our goal is to incorporate pixel-space supervision while avoiding backpropagation through the decoder $D$.

Given the predicted clean latent $\hat{\mathbf z}_0$, we decode an edge map $\hat{ y}$ and calculate the loss $\mathcal{L}_{\text{pix}}$ against the ground truth edge map $y$.
$D$ is used only to compute the forward value of $\mathcal{L}_{\text{pix}}$.

Rather than differentiating $\mathcal{L}_{\text{pix}}$ through $D$, we inject a proxy upstream gradient to $\hat{\mathbf z}_0$.
Concretely, we form the injected gradient as a broadcasted tensor $\mathbf g = \mathcal{L}_{\text{pix}}\,\mathbf 1$ with the same shape as $\hat{\mathbf z}_0$.
We then use a custom autograd operator that overrides the backward gradient on $\hat{\mathbf z}_0$ as
\begin{equation}
  \frac{\partial \mathcal{L}_{\text{pix}}}{\partial \hat{\mathbf z}_0} \leftarrow \mathbf g,
  \qquad
  \nabla_\theta \mathcal{L}_{\text{pix}} \approx
  \frac{\partial \hat{\mathbf z}_0}{\partial \theta}^{\!\top}\mathbf g .
\end{equation}
This avoids the memory and compute overhead of decoder backpropagation, while allowing pixel-space supervision to influence the latent dynamics.

We summarize the procedure in the following pseudocode.
\texttt{SpecifyGradient} is a custom autograd operator whose forward output is used only to connect the computation graph, while its backward pass overrides the gradient of the loss with respect to $\hat{\mathbf z}_0$ with the injected signal $\mathbf g$.

\paragraph{Pseudocode.}
\begin{lstlisting}[mathescape=true, basicstyle=\ttfamily\footnotesize, columns=fullflexible, frame=single, breaklines=true, showstringspaces=false, numbers=left, numberstyle=\tiny\ttfamily, numbersep=8pt, xleftmargin=2.2em, framexleftmargin=2.0em]
Inputs: predicted clean latent $\hat{\mathbf z}_0$, ground truth edge map $y$, decoder $D$
compute $\hat{y} \gets D(\hat{\mathbf z}_0)$
compute scalar loss $\mathcal{L}_{\text{pix}}$ by comparing $\hat{y}$ and $y$
compute $L_{\text{inj}} \gets \mathrm{SpecifyGradient}(\hat{\mathbf z}_0, \mathcal{L}_{\text{pix}})$
   # in backward, $\mathcal{L}_{\text{pix}}$ is broadcast to $\mathbf g = \mathcal{L}_{\text{pix}}\,\mathbf 1$
backprop $L_{\text{inj}}$    # d/d$\hat{\mathbf z}_0$ is overridden by $\mathbf g$
\end{lstlisting}






This procedure does not claim to compute the exact gradient of $\mathcal{L}_{\text{pix}}$ through $D$.
Instead, it injects a proxy gradient derived from $\mathcal{L}_{\text{pix}}$ as an upstream signal on $\hat{\mathbf z}_0$.
We expect it to improve accuracy in practice because the injected signal scales the update magnitude according to pixel-space errors, effectively reallocating optimization budget toward harder samples with larger $\mathcal{L}_{\text{pix}}$.
\section{Additional Experimental Results}
\label{sec:additional_results}

\subsection{Additional Qualitative Results}
\label{app:additional_qualitative_results}
Fig.~\ref{fig:appx_qual} presents additional qualitative results across BSDS500, NYUDv2, BIPED, and CubiCasa datasets.
Our method produces crisp edge maps without post-processing across diverse scene types, demonstrating the effectiveness of foundation-model adaptation for edge detection.
The results show that our approach captures fine-grained details and maintains sharp boundaries.

\subsection{Impact of Multiple Inference Steps}
\label{app:inference_steps}
\begin{figure}[t]
  \centering
  \includegraphics[width=0.8\linewidth]{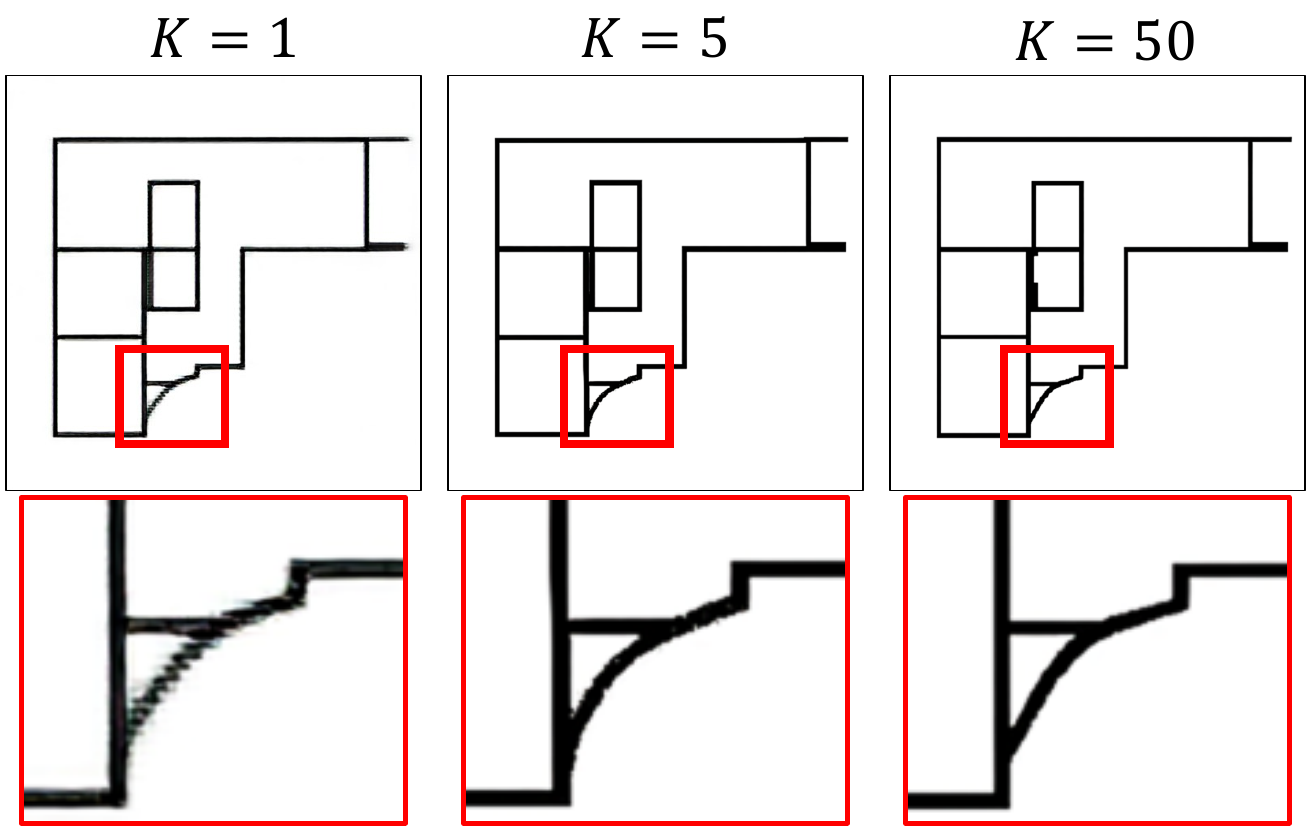}
  \caption{Qualitative comparison of edge detection results with varying inference step numbers $K$.
    Multi-step inference ($K > 1$) produces crisper and more refined edge maps.}
  \label{fig:stepnum_results}
\end{figure}
\begin{figure}[t]
  \centering
  \includegraphics[width=\columnwidth]{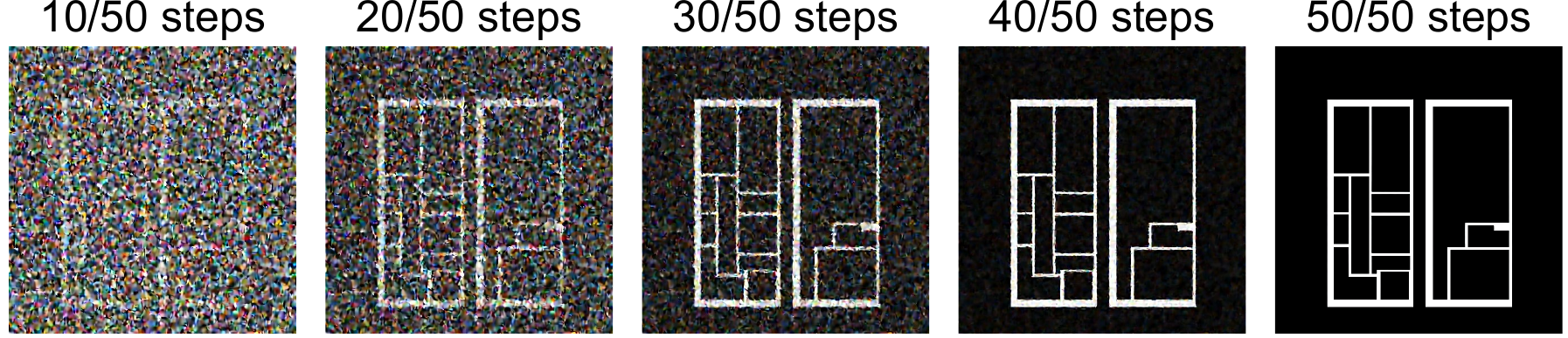}
  \caption{
    The generated edge maps of CubiCasa dataset from 10/50 to 50/50 steps.
    As the number of steps increases, boundaries stabilize.}
  \label{fig:generative_process}
\end{figure}
Fig.~\ref{fig:stepnum_results} presents qualitative results for different inference step numbers $K$.
These results demonstrate that multi-step inference ($K > 1$) achieves significantly crisper edge maps compared to single-step generation.

Fig.~\ref{fig:generative_process} also shows the generation trajectory of our model (10/50 to 50/50 steps).
With fewer steps, the global structure appears but boundaries remain unstable; as the number of steps increases, boundaries stabilize.
This indicates that multi-step refinement is important for obtaining crisp edges.

\vfill\pagebreak


\end{document}